\documentclass[letterpaper]{article} 
\usepackage{aaai24}  
\usepackage{times}  
\usepackage{helvet}  
\usepackage{courier}  
\usepackage[hyphens]{url}  
\usepackage{graphicx} 
\usepackage{natbib}  
\usepackage{caption} 
\usepackage{amsmath}
\usepackage{booktabs}
\usepackage{multirow}

\title{SAMFlow: Eliminating Any Fragmentation in Optical Flow with Segment Anything Model}
\author {
    Shili Zhou,
    Ruian He,
    Weimin Tan\thanks{Corresponding author: Weimin Tan and Bo Yan. This work is supported by NSFC (GrantNo.: U2001209 and 62372117) and Natural Science Foundation of Shanghai (21ZR1406600).},
    Bo Yan\footnotemark[1]
}
\affiliations {
    School of Computer Science, Shanghai Key Laboratory of Intelligent Information Processing, Fudan University\\
    slzhou19@fudan.edu.cn, rahe16@fudan.edu.cn, wmtan@fudan.edu.cn, byan@fudan.edu.cn
}

\begin{document}

\maketitle

\begin{abstract}
Optical Flow Estimation aims to find the 2D dense motion field between two frames. Due to the limitation of model structures and training datasets, existing methods often rely too much on local clues and ignore the integrity of objects, resulting in fragmented motion estimation. Through theoretical analysis, we find the pre-trained large vision models are helpful in optical flow estimation, and we notice that the recently famous Segment Anything Model (SAM) demonstrates a strong ability to segment complete objects, which is suitable for solving the fragmentation problem. We thus propose a solution to embed the frozen SAM image encoder into FlowFormer to enhance object perception. To address the challenge of in-depth utilizing SAM in non-segmentation tasks like optical flow estimation, we propose an Optical Flow Task-Specific Adaption scheme, including a Context Fusion Module to fuse the SAM encoder with the optical flow context encoder, and a Context Adaption Module to adapt the SAM features for optical flow task with Learned Task-Specific Embedding. Our proposed SAMFlow model reaches \textbf{0.86/2.10} clean/final EPE and \textbf{3.55/12.32} EPE/F1-all on Sintel and KITTI-15 training set, surpassing Flowformer by \textbf{8.5\%/9.9\%} and \textbf{13.2\%/16.3\%}. Furthermore, our model achieves state-of-the-art performance on the Sintel and KITTI-15 benchmarks, \textbf{ranking \#1} among all two-frame methods on Sintel clean pass.

\end{abstract}
\begin{figure}[h]
    \centering
    \includegraphics[width=\columnwidth]{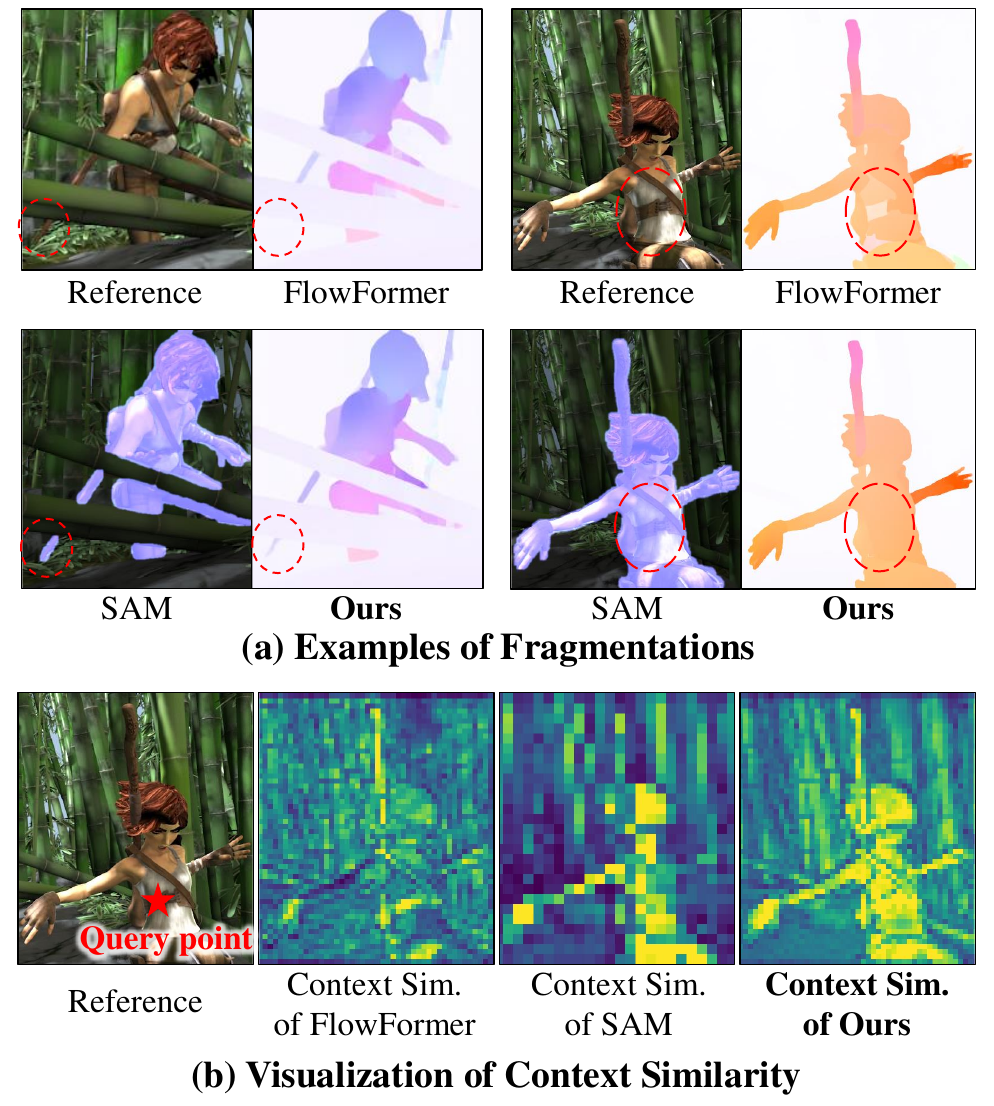}
    \caption{\textbf{(a)} Examples of \textbf{fragmentation} in optical flow estimation. We observe that SAM is able to segment the whole object. Thus, we propose our SAMFlow to eliminate fragmentation. (b)Visualization of the context similarity with the query point.} 
    \label{fig:motivation}
\end{figure}

\section{Introduction}
Optical flow is a fundamental task in computer vision, which estimates the pixel-level correspondences between frames. As an important paradigm to exploit video temporal continuity, it has applications in many video-related downstream tasks, such as frame interpolation \cite{RIFE}, video inpainting \cite{video_completion} and action recognition \cite{video_action}. With the advent of advanced neural network architectures, many powerful optical flow estimation models have been proposed \cite{dosovitskiy2015flownet,sun2018pwc,teed2020raft,jiang2021GMA,huang2022flowformer}.

Although leaps and bounds have been made, existing optical flow estimation methods are still limited by two factors: 1) \textbf{Scarcity of 
well-labeled datasets}. Since it is difficult to obtain pixel-level motion annotations in the real world, optical flow datasets are usually constructed using artificial synthesis schemes. For example, some works \cite{dosovitskiy2015flownet, sun2021autoflow} try to construct datasets from images and generate motions with simple 2D transformations. At the same time, other works \cite{mayer2016flyingthings, VirtualKITTI} generate datasets of virtual scenes with the 3D rendering engine. Compared with natural scenes, these synthetic datasets have limited diversity and realism, resulting in insufficient training of existing optical flow models. 2) \textbf{Lack of high-level understanding}. Human perception of motion is closely linked to the understanding of objects. Instead, optical flow models only focus on the local low-level clues, leading to incorrect ``\textbf{fragmentation}" results. Here, fragmentation refers to erroneous fragmented optical flow predictions for the same object. Figure \ref{fig:motivation} (a) give examples of fragmentation in optical flow caused by occlusion and complex lighting/textures. Some previous studies \cite{sun2022skflow, jiang2021GMA} also try to solve this problem by using larger receptive fields or global motion aggregation. However, these simple structural improvements cannot fundamentally eliminate fragmentation.

The recent pre-trained large vision models that have received considerable attention are highly suitable for addressing the aforementioned two challenges. (\romannumeral1) \cite{shi2023flowformer++} and \cite{dong2023rethinking} have shown that pre-training with data and supervision beyond optical flow can strengthen optical flow estimation, which implies that pre-trained large vision models can leverage a wide range of unlabeled image and video data to circumvent the problem of insufficient optical flow datasets. (\romannumeral2) The visual representation learned through pre-training contains the high-level understanding we need. Therefore, fusion with large vision models may further enhance optical flow estimation. Among the large vision models, Segment Anything Model (SAM) \cite{kirillov2023SAM} is one of the most suitable for optical flow estimation. As shown in Figure \ref{fig:motivation} (a), SAM can segment entire objects under occlusion and other confusing environments, which is exactly the solution to fragmentation in optical flow. Here, we propose using SAM's features as the SAM image encoder occupies most of the SAM parameters and knowledge.

However, it is challenging to effectively harness SAM features for non-segmentation tasks such as optical flow estimation due to the absence of task-specific knowledge. As illustrated in Figure \ref{fig:motivation}(b), while SAM's feature yields a superior similarity map compared to FlowFormer, it losses numerous details, posing an obstacle to optical flow estimation. Therefore, we propose an Optical Flow Task-Specific Adaptation scheme to address the challenge. First, we fuse the SAM encoder with the task-specific encoder with a Context Fusion Module (CFM). Next, we introduce a Context Adaption Module (CAM) to inject more task-specific knowledge of optical flow into the fused features via Two-Way Attention (TWA) blocks and Learned Task-Specific Embedding (LTSE) tokens. With the above designs, our proposed SAMFlow achieves remarkable performance, reaching \textbf{0.86/2.10} clean/final EPE on Sintel \cite{butler2012sintel} training set and \textbf{3.55/12.32} EPE/F1-all on KITTI-15 \cite{geiger2013kitti} training set, surpassing Flowformer by \textbf{8.5\%/9.9\%} and \textbf{13.2\%/16.3\%}. Furthermore, we upload our fine-tuned models to the benchmark sites of Sintel and KITTI-15, which shows significant superiority, \textbf{ranking \#1} among all two-frame methods on Sintel clean pass.

In summary, our contributions are as follows:
\begin{itemize}
\item For the first time, we investigate the feasibility of utilizing pre-trained SAM in optical flow estimation, and we thus propose SAMFlow, a novel approach aimed at enhancing the accuracy of optical flow estimation by effectively addressing issues of fragmentation.
\item To prevent the task mismatch from affecting the accuracy, we propose an Optical Flow Task-Specific Adaptation scheme by introducing the CFM to fuse the SAM encoder with the optical flow context encoder, and the CAM to further adapt for optical flow estimation, improving the effectiveness of SAMFlow significantly.
\item Our SAMFlow achieves state-of-the-art performance on both generalization and dataset-specific evaluations, surpassing Flowformer with a large margin and ranking \#1 among all two-frame methods on the clean pass of the Sintel benchmark. 
\end{itemize}

\section{Related Works}

\subsection{Optical Flow}
Optical flow has been studied for many years as a fundamental vision task. Traditional methods such as \cite{lucas198LK} and \cite{horn1981HS} regard optical flow estimation as an energy optimization task and use human-designed data and prior terms as optimization objectives, which cannot satisfy the complex motion in natural images. 
In recent years, benefiting from the emergence of deep learning and large-scale synthetic optical flow datasets, most high-performance optical flow estimation methods learn optical flow automatically in an end-to-end manner. Model design and data collection replace the data and prior term, and become the focus of today's optical flow algorithm researchers. 

For the model, the researchers successively introduced convolutional network as FlowNet \cite{dosovitskiy2015flownet}, multi-scale network as PWC-Net \cite{sun2018pwc}, recurrent network as RAFT \cite{teed2020raft}, Transformer as FlowFormer \cite{huang2022flowformer} and other structures, making the inherent learning ability of the optical flow model enhanced gradually. Meanwhile, reducing model time-consuming is also a research direction like \cite{cheng2023contextaware}.

For data, from Chairs \cite{dosovitskiy2015flownet}, Things \cite{mayer2016flyingthings} to later AutoFlow \cite{sun2021autoflow}, Spring \cite{mehl2023spring}, the diversity and authenticity of synthetic datasets are increasing, while the richness of real datasets is also slowly increasing (KITTI \cite{geiger2013kitti} and HD1K \cite{kondermann2016hd1k}). 

These efforts open up the possibility of increasingly powerful optical flow estimation models. However, the scarcity of data and the limitation of model design are still the core problems of optical flow.

\begin{figure*}
    \centering
    \includegraphics[width=\textwidth]{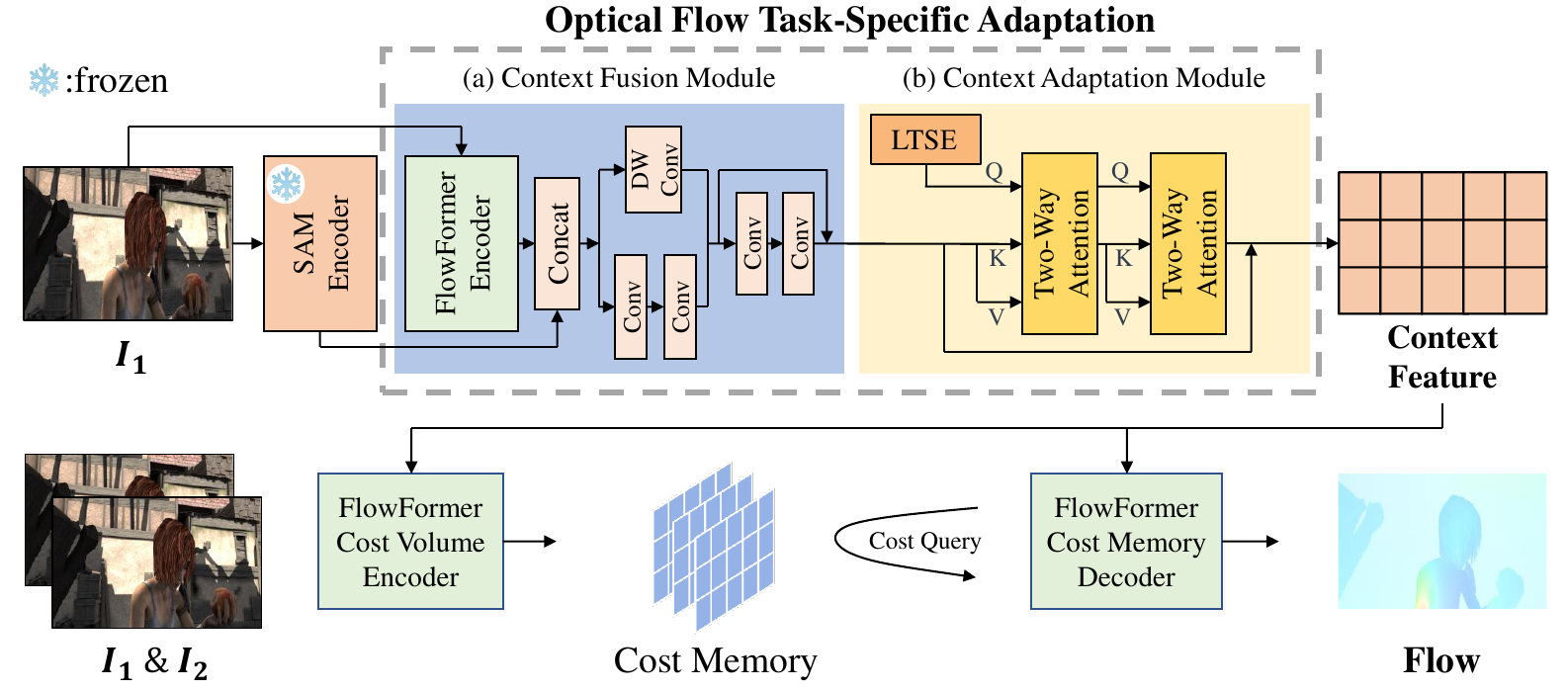}
    \caption{The overview of our SAMFlow, which utilizes the frozen SAM image encoder to boost the object perception of the optical flow model FlowFormer. We design two modules for in-depth utilizing SAM, including: (a) the CFM, which fuses SAM features with FlowFormer encoder, and (b) the CAM, which adapts the features with the Learned Task-Specific Embedding.}
    \label{fig:method}
\end{figure*}

\subsection{Pre-trained Large Vision Model}
The introduction to the pre-trained large vision model can be divided into the model architectures and the pre-training methods. We start by introducing the model architectures. In the early years, researchers use convolutional neural networks (CNN) as the basic architecture of computer vision, proposing VGG \cite{simonyan2014vgg}, ResNet \cite{he2016resnet}, \textit{etc}. Recently, inspired by the success of Transformer in natural language processing, Vision Transformer (ViT) \cite{dosovitskiy2020ViT} is proposed, which has a stronger representational ability and can show obvious advantages under large-scale datasets. 

Next, we introduce the pre-training methods. The early pre-training models using labeled data of pretext tasks, such as classification on ImageNet \cite{krizhevsky2012imagenet}. To use large-scale unlabeled data, researchers propose self-supervised pre-training methods, including contrastive learning \cite{chen2020contrastive}, auto-encoding \cite{vincent2008autoencoding}, \textit{etc}. A recent highlight paper \cite{he2022MAE} proposes Masked Auto-Encoder (MAE), which improves the traditional auto-encoder by dropping some patches when encoding to force the model to understand the image content.

\subsection{Segment Anything Model}
Segment Anything Model (SAM) \cite{kirillov2023SAM} is a prompt-based segmentation model. The structure of SAM is divided into three parts: image encoder, prompt encoder, and decoder. The image encoder is a variant of ViT \cite{dosovitskiy2020ViT}, which has a large number of parameters, while the prompt encoder and the decoder are lightweight. SAM is fine-tuned from MAE with a large amount of labeled segmentation data. This work also presents an impressive training data generation scheme: it uses manual labeling/correction and model learning/prediction as two complementary processes, which can create billions of segmentation labels with low labor costs. Large-scale labeled training data endows SAM with robust understanding and segmentation capabilities, which we find suitable for eliminating fragmentation in optical flow estimation.

\section{Proposed Method}

\subsection{Theoretical Analysis}
Optical flow estimation methods aim to find the mapping $\zeta:(I_1, I_2) \Rightarrow F$, where $I_1$ and $I_2$ are two adjacent frames from a video, and $F$ is the 2D optical flow field.
From a probabilistic point of view, an optical flow network can be expressed as:
\begin{equation}
F^* = \zeta_\theta(I_1, I_2) = \underset{F}{\operatorname{argmax}}\ {p(F \mid I_1, I_2)}
\end{equation}
where $F^*$ is the estimated most likely optical flow, $\zeta_\theta$ is the optical flow network with parameters $\theta$, and $p(F \mid I_1,I_2)$ is the posterior distribution of optical flow.

In order to facilitate a comprehensive analysis, we use Bayes' theorem to extend $p(F\mid I_1, I_2)$:
\begin{align}
p(F \mid I_1,I_2) &= \frac{p(F)p(I_1 \mid F)p(I_2 \mid I_1,F)}{p(I_1, I_2)} \\ \nonumber
&= \frac{p(I_1)p(F \mid I_1)p(I_2 \mid I_1,F)}{p(I_1, I_2)}
\end{align}

To find the optimal $F$, we omit the unrelated term $p(I_1)$ and $p(I_1, I_2)$, and take the logarithm to separate the multiplication terms. Thus, we get another formation of $F^*$ as in Formula \ref{eq:Fstar}, which consists of a \textbf{cost query term} and a \textbf{context term}.
\begin{align}
    F^* = \underset{F}{\operatorname{argmax}}\{\underbrace{\log p(I_2 \mid I_1, F)}_{\text {cost query}} + \underbrace{\log p\left(F\mid I_1\right)}_{\text {context}} \}
    \label{eq:Fstar}
\end{align}

The cost query term encompasses the interrelation between $I_2$ and $I_1$ with $F$. To achieve this, optical flow models such as \cite{sun2018pwc,teed2020raft,jiang2021GMA} construct the 4D cost volumes and make cost queries with flow guidance.

The contextual term provides an alternative information source for optical flow estimation, requiring the model to comprehend the image context more deeply. Earlier approaches like \cite{teed2020raft,jiang2021GMA,sun2022skflow} overlook this aspect, with the extracted contextual features being constrained to local cues. In contrast, our endeavor involves integrating pre-trained large vision models to enhance higher-level understanding. However, not all pre-trained large vision models prove apt for optical flow estimation, as certain models exclusively capture global semantics at the cost of losing spatial detail features, resulting in limited contributions to optical flow. Empirically, SAM is a suitable candidate with its capability to generate pixel-level outputs akin to optical flow, as shown in Figure \ref{fig:motivation}.

\subsection{Overview}
As shown in Figure \ref{fig:method}, we redesign the context feature extraction process of the backbone model FlowFormer by utilizing the image encoder of SAM, which has powerful object perception to solve the fragmentation of optical flow estimation. We call the proposed new model as SAMFlow. Moreover, as SAM does not acquire task-specific prior knowledge related to optical flow, we design an Optical Flow Task-Specific Adaption scheme, which includes a Context Fusion Module and a Context Adaption Module. In the following two subsections, we first introduce some minor modifications to unlock the resolution requirement of the SAM encoder; then, we introduce the CFM and CAM in detail.

\subsection{Modifications for Resolution}
The image encoder of SAM is a ViT that only accepts a fixed input resolution of $1024 \times 1024$. Considering the memory and time cost, this resolution is unable to use in the optical flow training framework. Therefore, we slightly modify the SAM encoder to unlock the resolution limitation. The details are provided in our supplementary material.

\subsection{Optical Flow Task-Specific Adaption}
\textbf{Context Fusion Module:}
Utilizing pre-trained large vision models for dissimilar tasks faces the challenge of knowledge unmatching. For example, the local details are essential clues for optical flow, while they are dropped for understanding tasks. Thus, we propose the CFM to combine the high-level understanding of SAM and the low-level clues for optical flow by using the SAM encoder and FlowFormer encoder simultaneously. 

As shown in Figure \ref{fig:method}(a), we first concatenate the SAM and Flowformer features. Subsequently, we mix them with two residual convolutional blocks. In the former block, the features will be processed by two branches: the main branch contains two 3x3 convolutional layers, which fuse the features and reduce channels, and the other branch uses depth-wise convolution directly to reduce the channels number and keep it consistent to the main branch. We add the results of the two branches as the output of the block. The latter residual block has almost the same structure except for depth-wise convolution since there is no difference in channel numbers between the input and output. Overall, this module can be represented by Formula \ref{eq:feat}, \ref{eq:cat}, \ref{eq:C0} and \ref{eq:C1}.
\begin{equation}
    \Phi_{S} = E_S(I), \Phi_{F} = E_F(I)
    \label {eq:feat}
\end{equation}
\begin{equation}
    \Phi_{\parallel} = \Phi_{S}\parallel \Phi_{F}
    \label {eq:cat}
\end{equation}
\begin{equation}
    \bar{\Phi}_C = Conv_2(Conv_1(\Phi_{\parallel})) + \Delta(\Phi_{\parallel})
    \label {eq:C0}
\end{equation}
\begin{equation}
    \Phi_C = Conv_4(Conv_3(\Phi_{\parallel})) + \bar{\Phi}_C
    \label {eq:C1}
\end{equation}
where $E_S$ and $E_F$ are the SAM encoder and FlowFormer Encoder, $\Phi_{S}$ and $\Phi_{F}$ are the extracted features, $\parallel$ is concatenation operator, $Conv_{k}$ corresponds to the k-th convolution layer. $\Delta$ is depth-wise convolution. $\Phi_{\parallel}$ and $\bar{\Phi}_C$ are intermediate variables, and $\Phi_C$ is the output of the CFM. The normalization and activation layers are omitted for brevity.

\begin{figure}
    \centering
    \includegraphics[width=\columnwidth]{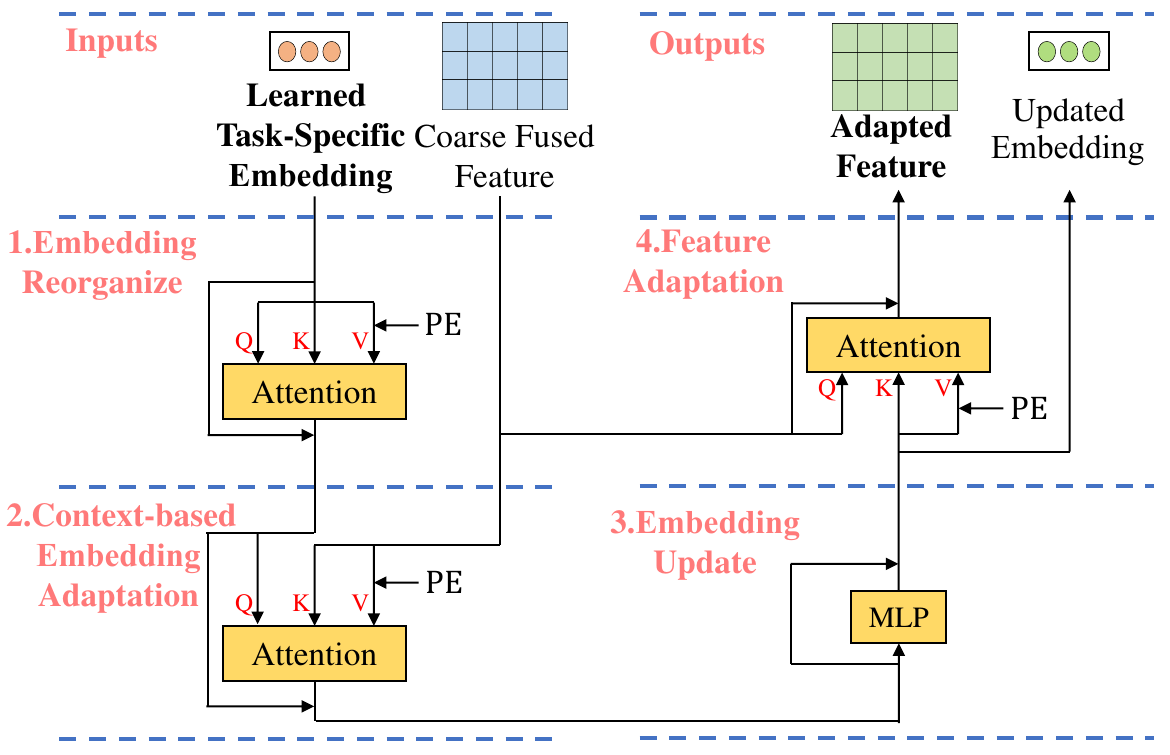}
    \caption{Our Context Adaptation Module to adapt SAM features for optical flow with the Learned Task-Specific Embedding. For the sake of brevity, only one Two-Way Attention is shown. PE is the positional embedding.}
    \label{fig:CAM}
\end{figure}

\noindent\textbf{Context Adaption Module:}
To better utilize the task-specific knowledge to accomplish task adaptation of optical flow, we propose the Context Adaption Module, as shown in Figure \ref{fig:method}(b) and \ref{fig:CAM}. Inspired by Perceiver IO \cite{jaegle2021perceiver} and the mask decoder of SAM, we make the following design in the Context Adaption Module: we use Learned Task-Specific Embedding (LTSE) tokens to store some task-specific priors of optical flow, and use Two-Way Attention (TWA) blocks to inject those priors into the context feature for adaptation.

The LTSE is implemented as a set of learnable offsets of shape $K\times D$, which will be automatically optimized during the training process. We empirically set $K$ to 3 and $D$ to 256. Meanwhile, each TWA contains four steps: 

1) \textbf{Embedding Reorganize}: as shown in Formula \ref{eq:TWA1}, a self-attention layer is used to reorganize the embedding of optical flow estimation task $\Omega_{T}$, which is the LTSE for the first TWA block. 
\begin{equation}
    \overline{\Omega}_{T} = \Omega_{T} + Att_1(\Omega_{T}, \Omega_{T}, \Omega_{T}+PE)
    \label {eq:TWA1}
\end{equation}
where $PE$ is the positional embedding. $Att_1$ is the first attention layer, which requires query, key, and value to be fed in order. We omit the normalization and activation layers here for brevity and do not expand the attention layer in detail. $\overline{\Omega}_{T}$ is the intermediate result of this step.

2) \textbf{Context-based Embedding Adaption}: as shown in Formula \ref{eq:TWA2}, we use a cross-attention layer to adapt the embedding with the context feature for better handling the input cases. 
\begin{equation}
    \hat{\Omega}_T = \bar{\Omega}_T + Att_2(\bar{\Omega}_T, \Phi_C, \Phi_C+PE)
    \label {eq:TWA2}
\end{equation}
where $\hat{\Omega}$ is the adaptation result of this step.

3) \textbf{Embedding Update}: as shown in Formula \ref{eq:TWA3}, we use a Multi-layer Perceptron (MLP) to update the query. 
\begin{equation}
    \Omega_U = \hat{\Omega}_T + MLP(\hat{\Omega}_T)
    \label {eq:TWA3}
\end{equation}
where $MLP$ is the Multi-layer Perceptron, and $\Omega_U$ is the updated embedding.

4) \textbf{Feature Adaption}: we use the updated embedding to adapt the context feature for optical flow tasks with a cross-attention layer, as shown in Formula \ref{eq:TWA4}.
\begin{equation}
    \Phi_C^A = \Phi_C + Att_3(\Phi_C, \Omega_U, \Omega_U+PE)
    \label {eq:TWA4}
\end{equation}
where $\Phi_C^A$ is the adapted context feature under the guidance of optical flow task-specific queries.

Finally, we use an addition operation to blend the results of the two modules.

\begin{table}[t]
\centering
\begin{tabular}{@{}clcccc@{}}
\toprule
\multirow{2}{*}{\begin{tabular}[c]{@{}c@{}}Training\\ Stage\end{tabular}} & \multirow{2}{*}{Method} & \multicolumn{2}{l}{Sintel(train)} & \multicolumn{2}{l}{KITTI-15(train)} \\ \cmidrule(l){3-6} 
                       &                         & clean           & final           & EPE              & F1               \\ \midrule
\multirow{15}{*}{C+T}  & HD3                     & 3.84            & 8.77            & 13.17            & 24.0             \\
                       & LiteFlowNet             & 2.48            & 4.04            & 10.39            & 28.5             \\
                       & PWC-Net                 & 2.55            & 3.93            & 10.35            & 33.7             \\
                       & LiteFlowNet2            & 2.24            & 3.78            & 8.97             & 25.9             \\
                       & S-Flow                  & 1.30            & 2.59            & 4.60             & 15.9             \\
                       & RAFT                    & 1.43            & 2.71            & 5.04             & 17.4             \\
                       & FM-RAFT                 & 1.29            & 2.95            & 6.80             & 19.3             \\
                       & GMA                     & 1.30            & 2.74            & 4.69             & 17.1             \\
                       & GMFlow                  & 1.08            & 2.48            & -                & -                \\
                       & GMFlowNet               & 1.14            & 2.71            & 4.24             & 15.4             \\
                       & CRAFT                   & 1.27            & 2.79            & 4.88             & 17.5             \\
                       & SKFlow                  & 1.22            & 2.46            & 4.47             & 15.5             \\
                       & FlowFormer              & 0.94            & 2.33            & 4.09             & 14.72            \\
                       & FlowFormer++            & 0.90            & 2.30            & 3.93             & 14.13            \\
                       & Ours                    & \textbf{0.87}   & \textbf{2.11}   & \textbf{3.44}    & \textbf{12.28}   \\ \bottomrule
\end{tabular}
\caption{Generalization performance evaluation on Sintel and KITTI-15 train sets.}
\label{tab:sota_C+T}
\end{table}

\begin{table}[t]
\centering
\begin{tabular}{@{}clccc@{}}
\toprule
\multirow{2}{*}{\begin{tabular}[c]{@{}c@{}}Training\\ Stage\end{tabular}} & \multirow{2}{*}{Method} & \multicolumn{2}{c}{Sintel(test)} & KITTI-15(test) \\ \cmidrule(l){3-5} 
                                                                          &                         & clean           & final          & F1-all         \\ \midrule
\multirow{16}{*}{\begin{tabular}[c]{@{}c@{}}C+T+S\\ +K+H\end{tabular}}    & PWC-Net+                & 3.45            & 4.60           & 7.72           \\
                                                                          & VCN                     & 2.81            & 4.40           & 6.30           \\
                                                                          & MaskFlowNet             & 2.52            & 4.17           & 6.10           \\
                                                                          & S-Flow                  & 1.50            & 2.67           & 4.64           \\
                                                                          & RAFT                    & 1.94            & 3.18           & 5.10           \\
                                                                          & RAFT*                   & 1.61            & 2.86           & 5.10           \\
                                                                          & FM-RAFT                 & 1.72            & 3.60           & 6.17           \\
                                                                          & GMA                     & 1.40            & 2.88           & 5.15           \\
                                                                          & GMA*                    & 1.39            & 2.47           & 5.15           \\
                                                                          & GMFlow                  & 1.74            & 2.90           & 9.32           \\
                                                                          & GMFlowNet               & 1.39            & 2.65           & 4.79           \\
                                                                          & CRAFT                   & 1.45            & 2.42           & 4.79           \\
                                                                          & SKFlow*                 & 1.28            & 2.23           & 4.84           \\
                                                                          & FlowFormer              & 1.16            & 2.09           & 4.68           \\
                                                                          & FlowFormer++            & 1.07            & \textbf{1.94}  & 4.52           \\
                                                                          & Ours                    & \textbf{1.00}   & 2.08           & \textbf{4.49}  \\ \bottomrule
\end{tabular}
\caption{Benchmark evaluation on Sintel and KITTI-15 test sets. The models with * adopt the warm-start strategy proposed in \cite{teed2020raft}.}
\label{tab:sota_C+T+S+K+H}
\end{table}

\section{Experiment}

\subsection{Settings}

\textbf{Training Settings} We follow the setup of previous work \cite{huang2022flowformer} and divide the training into two stages: C+T-Stage and C+T+S+K+H-stage. To speed up training, we skip the stage of training on the Chairs dataset by using FlowFormer-things checkpoint as initialization, and the SAM encoder is kept frozen during training.

\noindent \textbf{Test Settings} For testing, we adopt the tiling strategy \cite{jaegle2021perceiver} to bridge the resolution gap between training and testing data.

\begin{table}[t]
\centering
\begin{tabular}{@{}lcccc@{}}
\toprule
\multirow{2}{*}{Methods} & \multicolumn{2}{c}{Sintel (train) Occ.} & \multicolumn{2}{c}{Sintel (test) Occ.} \\ \cmidrule(l){2-5} 
                         & clean              & final             & clean             & final             \\ \midrule
RAFT                     & 5.36               & 7.09              & 9.65              & 14.68             \\
GMA                      & 4.25               & 6.22              & 7.96              & 12.50             \\
SKFlow                   & 3.44               & 4.52              & 7.25              & 11.42             \\
FlowFormer               & 2.76               & 3.60              & 7.16              & 11.30             \\
FlowFormer++             & 2.54               & 3.41              & 6.64              & 10.63             \\
Ours           & \textbf{2.24}      & \textbf{2.99}     & \textbf{5.97}     & \textbf{10.60}    \\ \bottomrule
\end{tabular}
\caption{Evaluation in occluded area of Sintel train and test sets.}
\label{tab:occlusion}
\end{table}
\begin{figure}[t]
    \centering
    \includegraphics[width=\columnwidth]{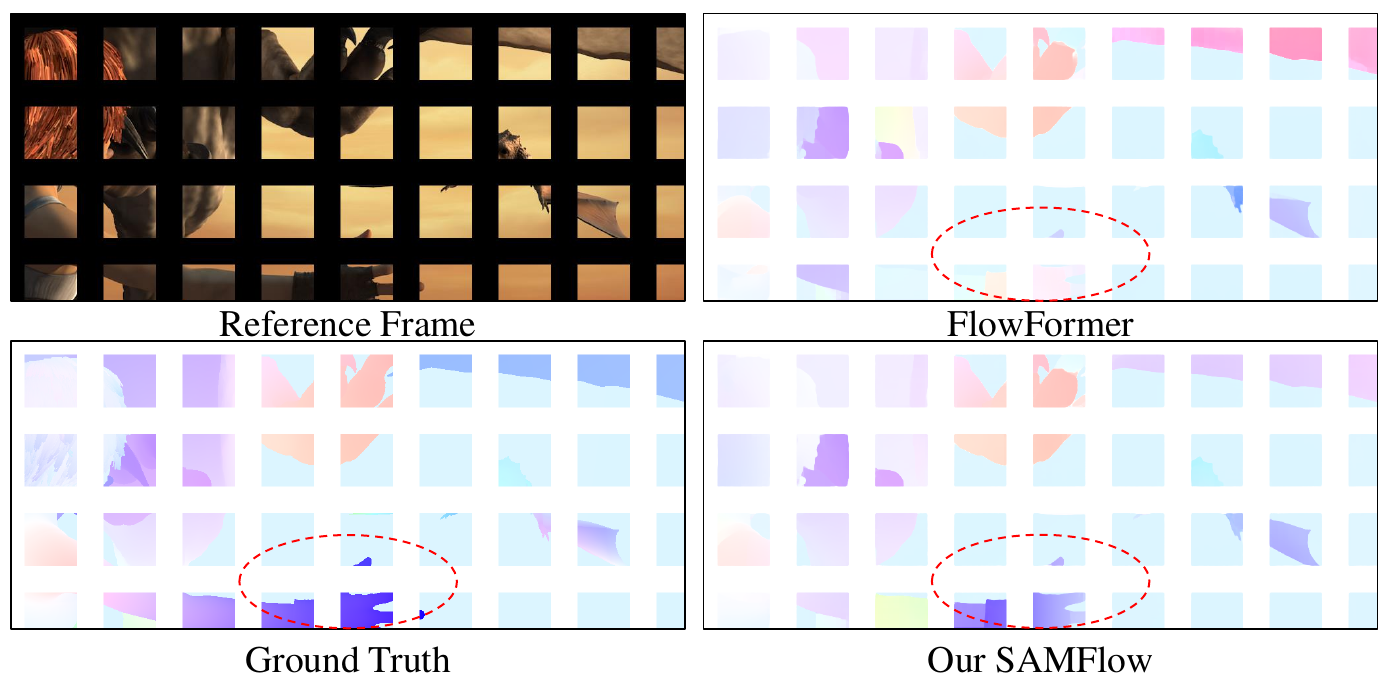}
    \caption{An example of the fragmentation attack, where our SAMFlow shows robustness over FlowFormer.}
    \label{fig:attack}
\end{figure}
\begin{figure}
    \centering
    \includegraphics[width=\columnwidth]{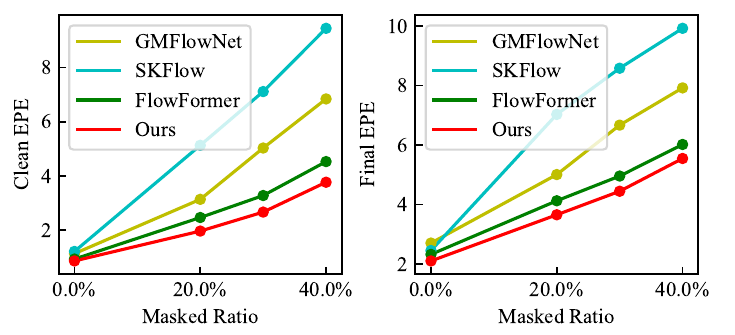}
    \caption{The average EPE of sintel clean and final pass under the fragementation attack with different masked ratios.}
    \label{fig:attack1}
\end{figure}

\subsection{Quantitative Comparison}
We first use the model trained in the C+T-stage for evaluating the generalization performance on the training sets of Sintel and KITTI. Then, we uploaded the results of the C+T+S+K+H-stage model and the K-stage model to the Sintel Benchmark website and the KITTI Benchmark website to compare the dataset-specific accuracy with the SOTA methods, including HD3 \cite{yin2019HD3},LiteFlowNet \cite{hui2018liteflownet}, PWC-Net \cite{sun2018pwc}, PWC-Net++ \cite{sun2019PWC++}, LiteFlowNet2 \cite{hui2020liteflownet2}, S-Flow \cite{zhang2021separable}, RAFT \cite{teed2020raft}, FM-RAFT \cite{jiang2021FMRAFT}, GMA \cite{jiang2021GMA}, GMFlow \cite{xu2022gmflow}, GMFlowNet \cite{zhao2022gmflownet}, CRAFT \cite{sui2022craft}, SKFlow \cite{sun2022skflow}, FlowFormer \cite{huang2022flowformer} and FlowFormer++ \cite{shi2023flowformer++}.

\noindent \textbf{Generalization Performance}
As shown in Table \ref{tab:sota_C+T}, for the C+T-stage, our model achieves the best performance on all metrics on the training set of Sintel and KITTI-15 datasets. The EPEs of our SAMFlow on Sintel clean and final pass reach \textbf{0.86} and \textbf{2.10}. SAMFlow also achieve \textbf{3.55} EPE and \textbf{12.32} F1 on KITTI-15 datasets. It is worth noting that FlowFormer uses two different model checkpoints with different training patch-size to obtain better performance on Sintel and KITTI. In contrast, our method uses the same checkpoint when evaluating both datasets. Nevertheless, our SAMFlow still easily surpasses the performance of FlowFormer, reducing Sintel clean/final EPE and KITTI-15 EPE/F1 by \textbf{8.5\%}/\textbf{9.9\%} and \textbf{13.2\%}/\textbf{16.3\%}, respectively.

\noindent \textbf{Comparison on Benchmarks}
Table \ref{tab:sota_C+T+S+K+H} also proves the dataset-specific performance of our SAMFlow. On the Sintel test set, our method achieves \textbf{1.00} clean EPE and \textbf{2.08} final EPE. Meanwhile, on the KITTI-15 test set, SAMFlow achieves \textbf{4.49} F1-all. Compared with FlowFormer, our method has achieved all-around improvement and can also defeat the new SOTA method FlowFormer++ on Sintel clean pass and KITTI-15. This demonstrates that our method brings significant accuracy improvements for optical flow estimation. The results can also be found on Sintel and KITTI-15 benchmark websites, where our SAMFlow \textbf{rank \#1} among all two-frame methods on Sintel clean pass.

\noindent \textbf{Evaluation in Occluded Area}
We give the comparison under occluded area, one of the significant sources of fragmentation, on the Sintel train set and the Sintel benchmark (test) as shown in Table \ref{tab:occlusion}. Compared with Flowformer, our method has improved by \textbf{18.84\%/17.94\%} and \textbf{16.62\%/6.19\%} on Sintel train and test, respectively. At the same time, it surpasses Flowformer++ on the Sintel benchmark and reaches the best.

\subsection{Fragmentation Attack}
To further demonstrate the ability of our method to eliminate fragmentation, we design the fragmentation attack, which splits the images into discrete parts using a grid-style mask, as shown in Figure \ref{fig:attack}. By controlling the thickness and density of the mask grid, we can mask out the images of the Sintel dataset at different ratios, creating different degrees of fragmentation. As shown in Figure \ref{fig:attack1}, we compare the robustness of GMFlow, SKFlow, FlowFormer, and our model under fragmentation attack with 0\%, 20\%, 30\% and 40\% masked ratios. Attacks greater than 40\% are meaningless because too much information has been lost, so we ignore these cases. It can be observed that SKFlow is greatly affected by fragmentation attacks. Its EPE increases sharply when adding 20\% mask. GMFlowNet and FlowFormer are also trapped in isolated local clues caused by fragmentation, showing a noticeable performance hit. With the object perception with SAM encoder, our method exhibits strong robustness by finding the relation between the image content in different grids, achieving better results than FlowFormer.

\begin{figure*}
    \centering
    \includegraphics[width=\textwidth]{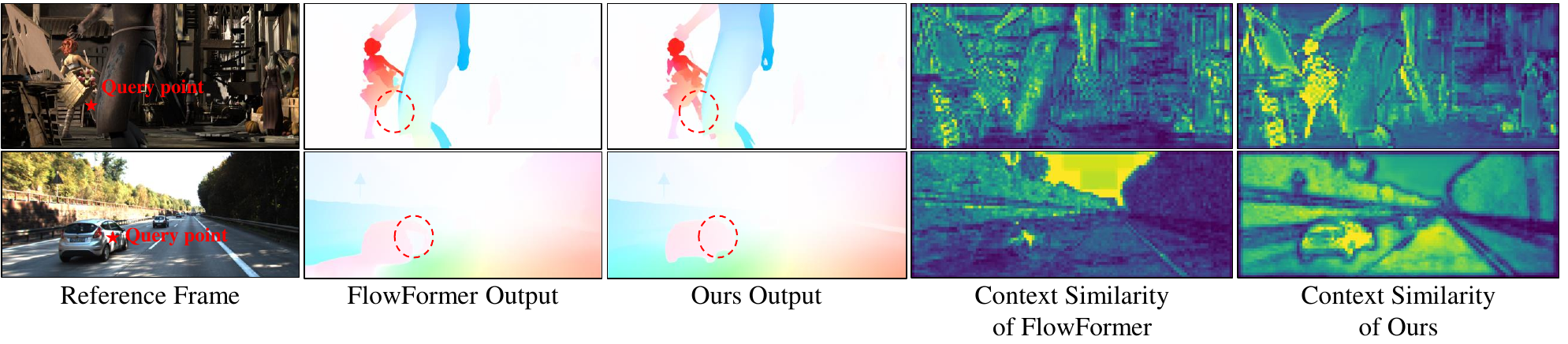}
    \caption{Two examples from Sintel and KITTI-15 for qualitative comparison corresponding to fragmentation caused by occlusion and complex lighting/textures, respectively.}
    \label{fig:compare}
\end{figure*}
\subsection{Visualization}
Figure \ref{fig:compare} shows two examples from Sintel and KITTI-15 for qualitative comparison, corresponding to fragmentation caused by occlusion and complex lighting/textures, respectively. We visualize the optical flow fields by mapping them to the color space and the context features by computing the feature similarity between all pixels and the chosen query points. We can find that FlowFormer context similarity is unordered and cannot guarantee the integrity of moving objects, resulting in the missing leg of the girl in the first example and the holes of the car in the second example. With the perception of objects, our SAMFlow gives better context features with apparent related objects and boundaries, thus enhancing the accuracy of optical flow greatly.

\subsection{Ablation Study}
We conduct a series of ablation studies to validate our SAMFlow, and the results are shown in Table \ref{tab:ablation}.

\textbf{Encoders and Modules:}
We compare different context feature settings to prove the effectiveness of our designs. The baseline model is FlowFormer, which only has the optical flow task-specific encoder. We first try using the SAM encoder instead of the FlowFormer encoder and find it brings some performance improvements on Sintel. However, the effect on authentic images (KITTI-15) is limited due to the lack of priors for optical flow estimation. Subsequently, we add our CFM, which fuses the FlowFormer and SAM features with residual convolutional blocks, significantly reducing the errors on KITTI-15 datasets. We further add our CAM to inject more task-specific knowledge into the context feature, which boosts the optical flow accuracy and achieves the best performances for both datasets. A CAM-only model was also added to the experiments to illustrate that both modules are necessary.

\textbf{SAM Model Scale:}
We try the SAM image encoders of different SAM scales, including \emph{SAM-H}, \emph{SAM-B} and 
a tiny version \emph{MobileSAM} \cite{zhang2023mobileSAM}. The baseline model is also listed, named \emph{w/o. SAM}. All our models outperform the baseline (FlowFormer), which once again proves the effectiveness of our proposed method. Moreover, we find that larger encoders show better results in general. However, there is an exception on the final pass of the Sintel dataset, where the MobileSAM encoder performs better than SAM-B encoder. This may be due to the different architectures of MobileSAM and SAM-B encoders, which cause them to behave differently under some specific scenes.

\textbf{Other Pre-trained Models} Besides SAM, we also try ViT \cite{dosovitskiy2020ViT}, MAE \cite{he2022MAE} and DINO \cite{zhang2022dino}. However, as shown in Table \ref{tab:ablation}, they do not work well. The reasons are two-fold: on the one hand, their tasks does not need good representations of spatial content; on the other hand, the pre-trained ViT and MAE suffer from a limitation of very low-resolution ($224\times 224$ or $384 \times 384$), making them poorly suited for larger inputs.

\begin{table}[]
\centering
\begin{tabular}{@{}lcccc@{}}
\toprule
\multirow{2}{*}{Methods} & \multicolumn{2}{c}{Sintel(train)} & \multicolumn{2}{c}{KITTI-15(train)} \\ \cmidrule(l){2-5} 
                         & clean           & final           & EPE              & F1               \\ \midrule
FlowFormer Enc.       & 0.94            & 2.33            & 4.09             & 14.72            \\
SAM Enc.              & 0.89            & 2.17            & 4.11             & 14.37            \\
CFM         & 0.89            & 2.11   & 3.83             & 13.43            \\
CAM         & 0.90            & \textbf{2.10}        & 3.81             & 13.23            \\
CFM + CAM      & \textbf{0.87}   & 2.11   & \textbf{3.44}    & \textbf{12.28}   \\ \midrule
w/o. SAM       & 0.94            & 2.33            & 4.09             & 14.72            \\
MobileSAM                & 0.88            & 2.19            & 3.78             & 13.22            \\
SAM-B                    & 0.88            & 2.26            & 3.57             & 12.45            \\
SAM-H                    & \textbf{0.87}   & \textbf{2.11}   & \textbf{3.44}    & \textbf{12.28}   \\
VIT                    & 1.08 & 2.38 & 4.01 & 13.47 \\
MAE                    & 1.01 & 2.35 & 3.91 & 13.56 \\
DINO                   & 0.94 & 2.26 & 4.07 & 13.76 \\
\bottomrule
\end{tabular}
\caption{Ablation study of encoder type, modules, and the scale of SAM encoder. We bold the best value in each group.}
\label{tab:ablation}
\end{table}

\begin{figure}
    \centering
    \includegraphics[width=0.9\columnwidth]{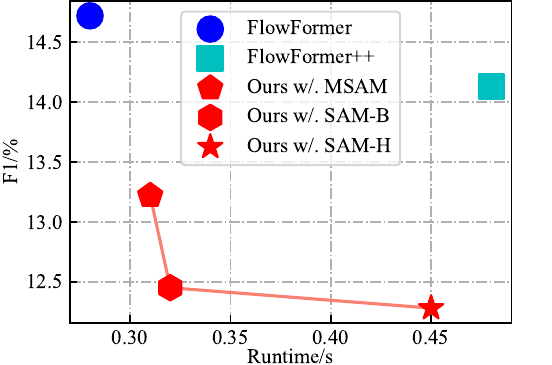}
    \caption{Runtime and accuracy comparison between Flowformer, FlowFormer++, and our models with different SAM encoders, including SAM-B, SAM-H, and MobileSAM (MSAM). The x-axis is the average time of 100 runs of $384\times1024$ inputs, and the y-axis is the f1 score on KITTI.}
    \label{fig:runtime}
\end{figure}

\subsection{Runtime Analysis}
There might be doubts about the computational cost of utilizing the SAM encoder. In order to analyze it, we compare the performance and runtime of our three models of different scales with FlowFormer and FlowFormer++, and the results are presented in Figure \ref{fig:runtime}. It can be found that our method can balance performance and runtime requirements by controlling the scale of the SAM encoder. Compared with FlowFormer, our SAMFlow w/. MSAM(MobileSAM) only slightly increases in runtime but shows a considerable drop in F1. Meanwhile, all of our three models are superior to FlowFormer++ in speed and performance, proving the practical significance of our method.

\section{Conclusion}
This paper focuses on the challenging fragmentation issues for optical flow estimation. We first give theoretical analysis for applying SAM feature in optical flow estimation. Thus, we propose SAMFlow, which incorporates SAM into the optical flow estimation network. Next, to address the unmatched task-specific knowledge between SAM and optical flow estimation, we introduce an Optical Flow Task-Speicifc Adaptation scheme, including the CFM and CAM. In experiments, we demonstrate the effectiveness of SAMFlow for fragmentation elimination and its superiority in terms of optical flow estimation accuracy, which achieves state-of-the-art performance, and ranks \#1 among all two-frame methods on Sintel clean pass.

\appendix
\begin{figure}
    \centering
    \includegraphics[width=\columnwidth]{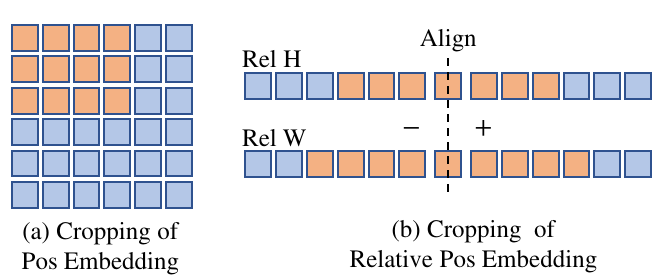}
    \caption{Cropping schemes for the Positional Encoding and the Relative Positional Encoding.}
    \label{fig:crop}
\end{figure}

\section{Details for Modifying SAM Encoder}
The SAM encoder requires a fixed input size $1024\times1024$, which is too large for the current optical flow training framework. By checking the code, we find that two parts limit the input resolution: the Positional Encoding (PE) and Relative Positional Encoding (RPE), implemented as fixed-size learnable offsets in the SAM encoder. We thus crop them to support smaller input sizes. On the one hand, we can assume that our input occupies the upper left corner of the full-size input, so we only need to crop the PE accordingly, as shown in Figure \ref{fig:crop}(a). On the other hand, RPE encodes both positive and negative offsets, using the midpoint of the RPE array to represent zero offsets. Thus, we cut off the same length at both ends as shown in Figure \ref{fig:crop}(b).

In addition, the downsampling factor of the SAM encoder is 16, while that of FlowFormer is 8. We add a bilinear interpolation upsampling layer and a convolutional layer after the SAM encoder to make them consistent.

\section{Details of Training Settings}
Our training proceeds as follows: In the first stage (i.e., the C+T-stage of previous work), we use Things dataset \cite{mayer2016flyingthings} to train for 120k steps, the learning rate is set to 1.25e-4, and the batch size is 3. The FlowFormer-things checkpoint is used as initialization. In the second stage (C+T+S+K+H-stage), we use a mixture of Things \cite{mayer2016flyingthings}, Sintel \cite{butler2012sintel}, KITTI-15 \cite{geiger2013kitti} and HD1K \cite{kondermann2016hd1k} datasets to fine-tune 240k steps. The learning rate is set to 5e-6, batch size is 3, and 2 gradient accumulations are used. In the third stage (K-stage), we continue to fine-tune 50k steps using KITTI-15 alone. The learning rate is set to 5e-6, and the batch size is set to 3. 

Three Nvidia RTX 3090 GPUs are used during the entire training process, where the SAM encoder is kept frozen. We use AdamW optimizer (weight\_decay=1e-5, epsilon=1e-8) and one cycle scheduler with linear decay strategy. Random crop (the patch size is 432$\times$960), random scale, and random flip are used as data augmentation. We adopt the mixed half-precision to speed up the training and add gradient clipping with a threshold of 1.0 for training stability.
\begin{figure*}[t]
    \centering
    \includegraphics[width=0.98\textwidth]{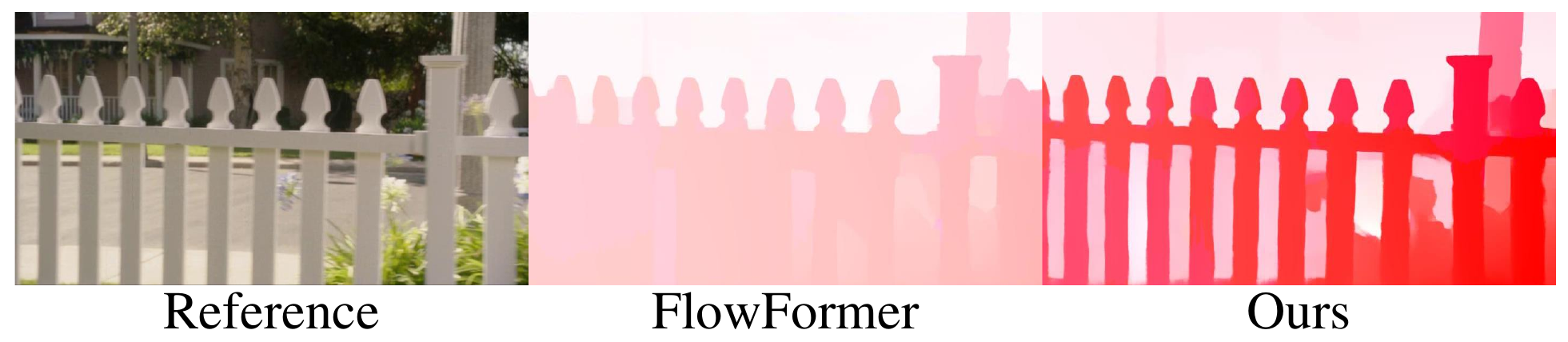}
    \caption{A real-world example of a challenging scenario. FlowFormer confuses the fence with the background, whereas our method is able to accurately distinguish.}
    \label{fig:re}
\end{figure*}

\begin{figure*}[t]
    \centering
    \includegraphics[width=\textwidth]{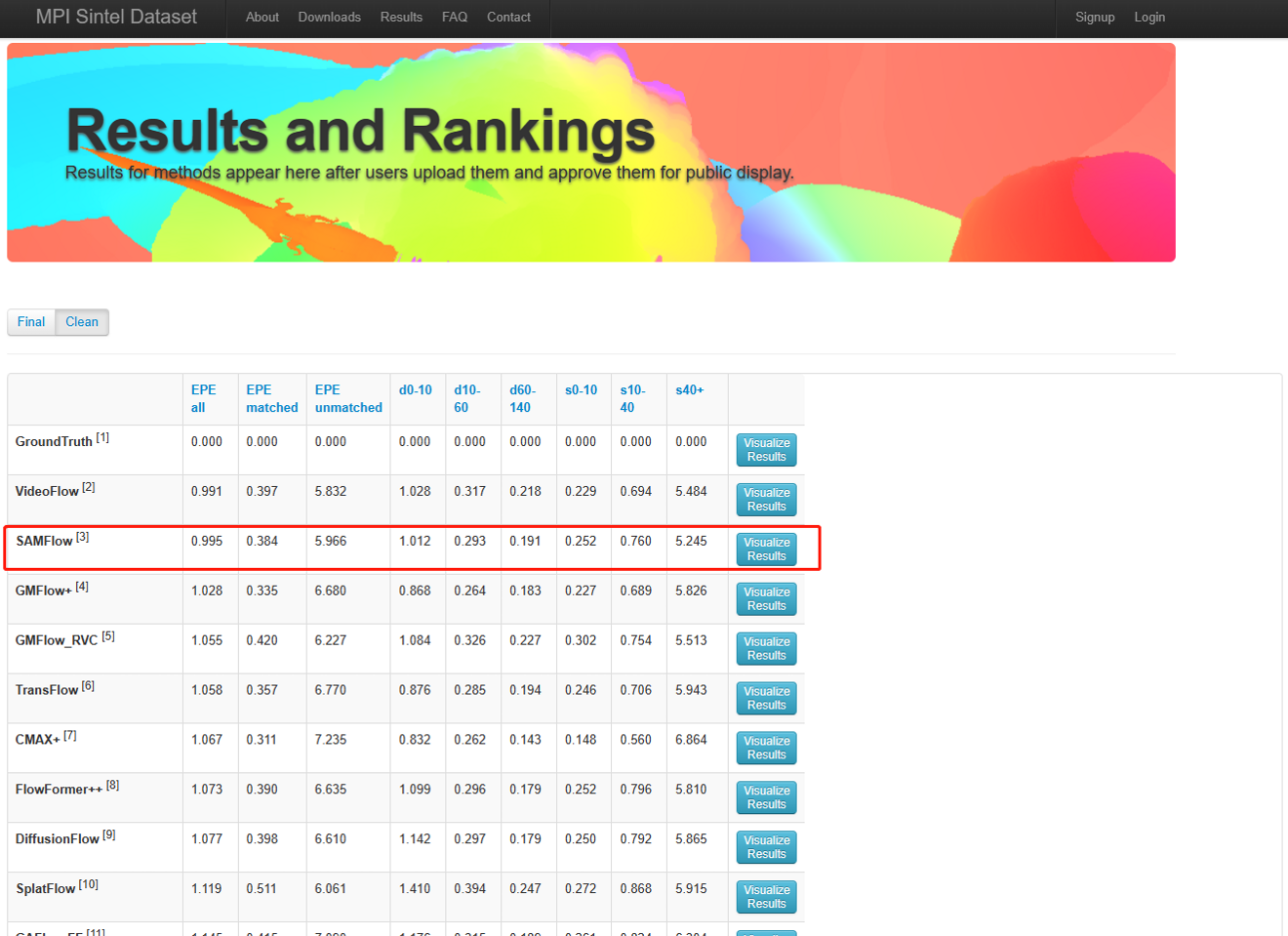}
    \caption{Screenshot of clean pass rankings on Sintel Benchmark.}
    \label{fig:screenshot}
\end{figure*}

\section{Testing on Real-World Scenarios}
We try using SAMFlow in complex real-world scenarios. An example is shown in the Figure \ref{fig:re}, where the fence is a faster-moving foreground, while objects such as pavement, flowers, and pillars form the background. Since there are a large number of hole areas in fences, it is easy to produce fragmented optical flow estimation results, causing the incorrect optical flow map estimated by FlowFormer in the figure. What's exciting is that our SAMFlow can correctly distinguish fences from other objects, and even correctly distinguish the layers of flowers, pavement, and pillars in the background.
\section{Screenshot on Sintel Benchmark Website}
We provide a screenshot of clean pass rankings on the MPI Sintel website \cite{butler2012sintel} in Figure \ref{fig:screenshot}. It can be found that our SAMFlow is just behind VideoFlow \cite{shi2023videoflow}, which utilizes five-frame inputs. In other words, our SAMFlow ranks \#1 among all two-frame methods.

\bibliography{mybib}
\end{document}